\PassOptionsToPackage{table}{xcolor}
\PassOptionsToPackage{normalem}{ulem}
\documentclass[]{academic_template}
\usepackage[toc,page,header]{appendix}

\usepackage{marvosym}

\usepackage{bbding}
\usepackage{soul}

\usepackage{amssymb}
\usepackage{amsmath}

\usepackage{makecell}
\usepackage{algorithm}      
\usepackage{algpseudocode}  
\usepackage{tabularx}
\usepackage{booktabs}
\usepackage{graphicx}
\usepackage{array}
\newcolumntype{C}{w{c}{3em}}
\usepackage{listings}
\usepackage{hyperref}
\usepackage{url}
\usepackage{booktabs}
\usepackage{wrapfig}

\usepackage{xcolor}
\usepackage{enumitem}
\definecolor{aceblue}{HTML}{3537EA} 

\usepackage{colortbl}
\usepackage{caption}
\usepackage{subcaption}

\usepackage{amsthm}

\def\eg{\emph{e.g.}}

\setlogoheight{8mm}   

\setlogospacing{0mm}
\setsjtublue 

\settitlerulethickness{3pt}  

\abstractboxon 
\setabstractframecolor{gray} 
\setabstractbgcolor{gray!10} 
\setlogotolineshift{0mm} 

\settoprulethickness{2.5pt}  
\setbottomrulethickness{1.5pt} 

\usepackage{dblfloatfix}
\usepackage{float}

\usepackage{tabularx}

\usepackage{multicol}
\usepackage{multirow}
\usepackage[table]{xcolor}  

\usepackage{natbib}
\setcitestyle{numbers}
\setcitestyle{square}
\usepackage{graphicx}

\usepackage{xspace}

\title{ACE-CaP-0: Real-World Code-as-Policies \\in the Era of Experience}

\author{%
\begin{minipage}{\textwidth}
    \centering
    \sffamily
    \setlength{\parskip}{0pt}

    {\fontsize{12}{16}\selectfont
    ACE-CaP Team\par}

    \vspace{0.65em}

    {\fontsize{10}{14}\selectfont
    Please see
    \hyperref[app:contributors]{Contributions and Author List}
    for more author details.\par}
\end{minipage}%
}

\abstract{
Recent advances in frontier models have pushed the boundaries of agentic robot control, enabling agents to perform complex robotic tasks through reasoning and interaction.
However, repeated calls to these models during execution can substantially increase task completion time.
Code as Policies (CaP) offers a route for coding agents to acquire physical skills through experience and retain them in executable programs, allowing the reasoning invested in development to be reused across runs.
Developing reliable code policies while limiting the cost of physical experimentation remains a central challenge.
We investigate human guidance as a source of priors for feasible actions and simulation as an auxiliary environment for exploring and testing candidate policies.
Our findings favor robot demonstrations available on request and simulation used selectively during real-world policy refinement.
These findings inform ACE-CaP-0, an agentic recipe that couples policy revision with experience acquisition through \textit{Demonstrations on Request} and \textit{Simulation on Demand}.
Across eight real-world manipulation tasks spanning alignment, multi-object handling, and bimanual coordination, the resulting frozen programs achieve 92.1\% success (221/240 trials).
Mean execution times across tasks range from 0.97 to 2.27 minutes.
The programs execute with runtime perception and control, without a human or coding agent in the loop.
These results highlight CaP's potential to turn experience into reliable, reusable physical skills.

}

\checkdata[Project Page]{\url{https://acerobotics2025.github.io/ACE-CaP/}}

\newif\ifusebibliography
\usebibliographyfalse

\begin{document}

\phantomsection
\addcontentsline{toc}{section}{Abstract}
\maketitle

\begin{figure}[H]
    \centering
    \includegraphics[width=\linewidth]{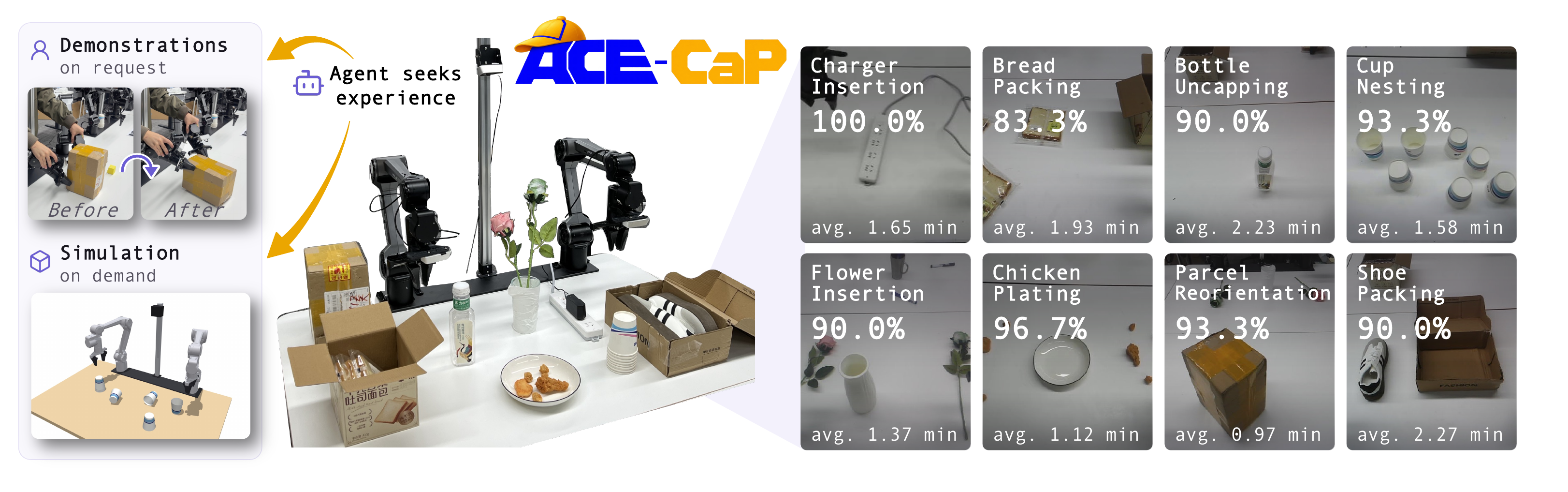}
    \caption{\textbf{ACE-CaP-0: from targeted experience to standalone robot execution.}
    Left: a coding agent refines policies on the robot, requesting robot demonstrations
    and using simulation as needed.
    Right: eight manipulation tasks, with each panel reporting the success rate
    and mean physical execution time in minutes.
    Each frozen policy is evaluated in 30 trials and executes with runtime
    perception and control, without a human or coding agent in the loop.}
    \label{fig:ace_cap_0_teaser}
\end{figure}

\newpage
\tableofcontents

\newpage
\section{Introduction}
\label{sec:introduction}

Foundation models are changing how robots acquire and execute skills.
Learned action models, including vision-language-action (VLA) and world action models (WAMs), combine pretrained knowledge with large-scale data and interaction experience to support generalist robot control~\cite{black2024pi_0,ye2026world,intelligence2025pi}.
Advances in frontier models (\eg, GPT-6 Astra~\cite{openai2026gpt6astra} and Claude Fable 5.2~\cite{anthropic2026fable52}) are expanding the role of agents in both robot control and policy development.
In agentic robot control, agents interpret observations, choose actions, and revise their decisions in response to physical feedback~\cite{jia2026agent}.
When execution relies on repeated model calls, however, inference latency can account for a substantial part of task completion time~\cite{sensenova2026gpt6astraachievedembodiedintelligence,ge2026make}.
Agents can also apply these capabilities during policy development, organizing experiments, diagnosing failures, and revising the policies a robot executes~\cite{xiao2026enpire,lu2026aspire}.

Code as Policies (CaP) represents robot behavior as executable programs, allowing decisions made during development to be retained and reused across runs~\cite{liang2023code,fu2026cap}.
Through demonstrations and experimentation, a coding agent can discover manipulation strategies and encode their action structure, geometric relationships, and motion parameters in code~\cite{kang2026encore}.
The resulting program can execute with runtime perception and control, without requiring the coding agent to select each action.
Its explicit task logic can be inspected and revised, supporting the accumulation of executable skills in embodied Recursive Self-Improvement (RSI)~\cite{hu2026embodied,lu2026aspire}.

Developing reliable code policies requires physical knowledge that a task instruction often leaves unspecified.
The agent must resolve choices about grasping, alignment, and coordination, yet a failed attempt may not reveal how to revise the program~\cite{kang2026encore,lu2026aspire}.
Repeated physical trials also consume robot time and can require human scene resets~\cite{xiao2026enpire,hu2026simex}.
Policy development therefore couples two decisions: how to revise the program, and what experience to acquire next.
We ask: \emph{how should agents acquire experience to develop reliable code policies when physical experimentation is costly?}

Human guidance can supply feasible strategies and motion references, while simulation provides an auxiliary environment for exploring and testing candidate implementations~\cite{kang2026encore,hu2026simex}.
We examine their use through six development conditions on charger insertion, with one development run per condition (Section~\ref{sec:investigation}).
\begin{itemize}[leftmargin=1.25em,itemsep=0.4em,topsep=0.4em,parsep=0pt]
    \item \textbf{What form of human guidance helps an agent develop a physically realizable policy, and when should it be provided?}
    Our results favor robot demonstrations paired with textual guidance and available on request during development.
    Robot demonstrations supply motion references in the target embodiment, while on-request access lets the agent seek guidance for its current manipulation problem.

    \item \textbf{When and how should simulated experience support real-world program development?}
    Our results favor using simulation on demand to investigate current development questions within a real-world feedback loop.
    Simulated experiments examine candidate revisions, while physical execution checks whether the sensing and action assumptions they rely on hold on the robot.
\end{itemize}

Building on these findings, we introduce \textbf{ACE-CaP-0}, a recipe that couples policy revision with experience acquisition through \emph{Demonstrations on Request} and \emph{Simulation on Demand} (Section~\ref{sec:recipe}).
During development, the coding agent chooses whether to request a robot demonstration, examine a candidate revision in simulation, or test the program on the robot.
The resulting program is frozen for evaluation and executes with runtime perception and control, without a human or coding agent in the loop.

We evaluate ACE-CaP-0 on eight real-world manipulation tasks requiring alignment, object reorientation, multi-object handling, and bimanual coordination (Section~\ref{sec:real_world_evaluation}).
Development allows at most ten human-assisted scene resets per task, excluding resets for demonstration collection.
Each frozen program is then evaluated in 30 trials without further updates.
As summarized in Figure~\ref{fig:ace_cap_0_teaser}, the programs achieve 92.1\% overall success (221/240 trials).
Mean physical execution times range from 0.97 to 2.27 minutes across tasks.
These results support CaP as a practical route from targeted experience to reliable, reusable physical skills.
\clearpage

\section{Investigation}
\label{sec:investigation}

We study charger insertion to examine how human guidance and simulation support code-policy development under limited physical interaction.
Six experimental conditions address three questions: what form of guidance to provide, when to provide it, and how to incorporate simulation into exploration.
Here, exploration involves executing candidate programs, inspecting feedback, and revising code.
All three comparisons share a real-only reference with on-request textual guidance and robot demonstrations.

\paragraph*{Shared setting and development budget.}
All conditions use the same foundation model, task instruction, and real-world perception and control interfaces.
We conduct one independent development run per condition, allowing at most ten human-assisted scene resets for the agent's physical trials.
Resets for demonstration collection are excluded; autonomous executions and demonstrations do not count against the budget.

\paragraph*{Evaluation and cost accounting.}
After development, we freeze each selected program and evaluate it in 30 real-world trials under common success criteria and initial-state conditions, without further updates~\cite{kress2024robot}.
Final success rate is the fraction of these trials in which the program completes charger insertion.
Development time is the elapsed time in hours, including model calls, simulated experiments, and real-world interaction.
Output token usage, reported in millions (M), sums the tokens generated across all model calls during development, including guidance interpretation, simulation construction, execution analysis, and code revision.
Both costs exclude frozen-policy evaluation.

\subsection{How Do Human Priors Shape Real-World CaP?}
\label{sec:investigation_human}

\begin{figure}[H]
    \centering
    \includegraphics[width=\linewidth]{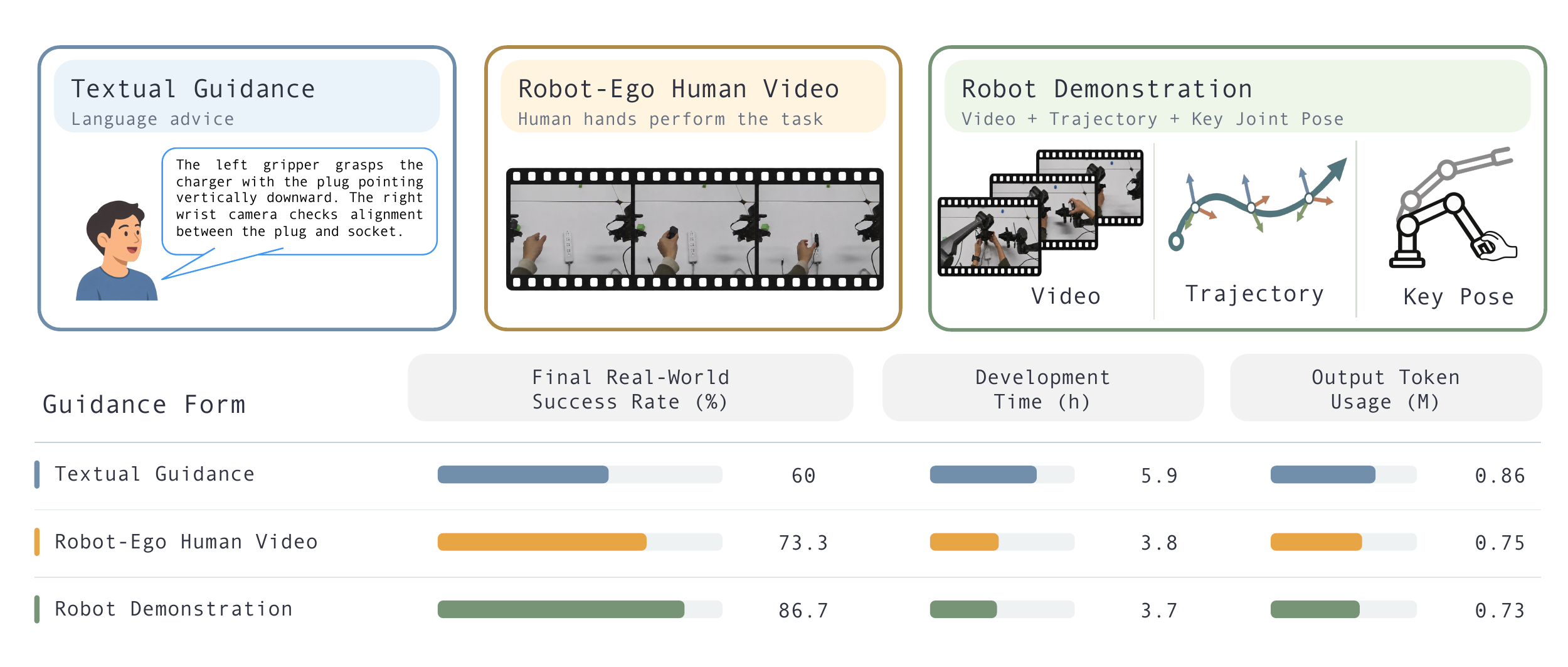}
    \caption{Guidance form for charger insertion. Top: textual guidance, robot-egocentric human video, and robot demonstrations, all available on request during real-only development. Both demonstration conditions also include textual guidance. Bottom: final success rate over 30 real-world trials of each frozen program, development time, and output token usage. Robot Demonstrations is the shared reference in Figures~\ref{fig:investigation_guidance_timing} and~\ref{fig:investigation_simulation}.}
    \label{fig:investigation_guidance}
\end{figure}

\begin{figure}[htbp]
    \centering
    \includegraphics[width=\linewidth]{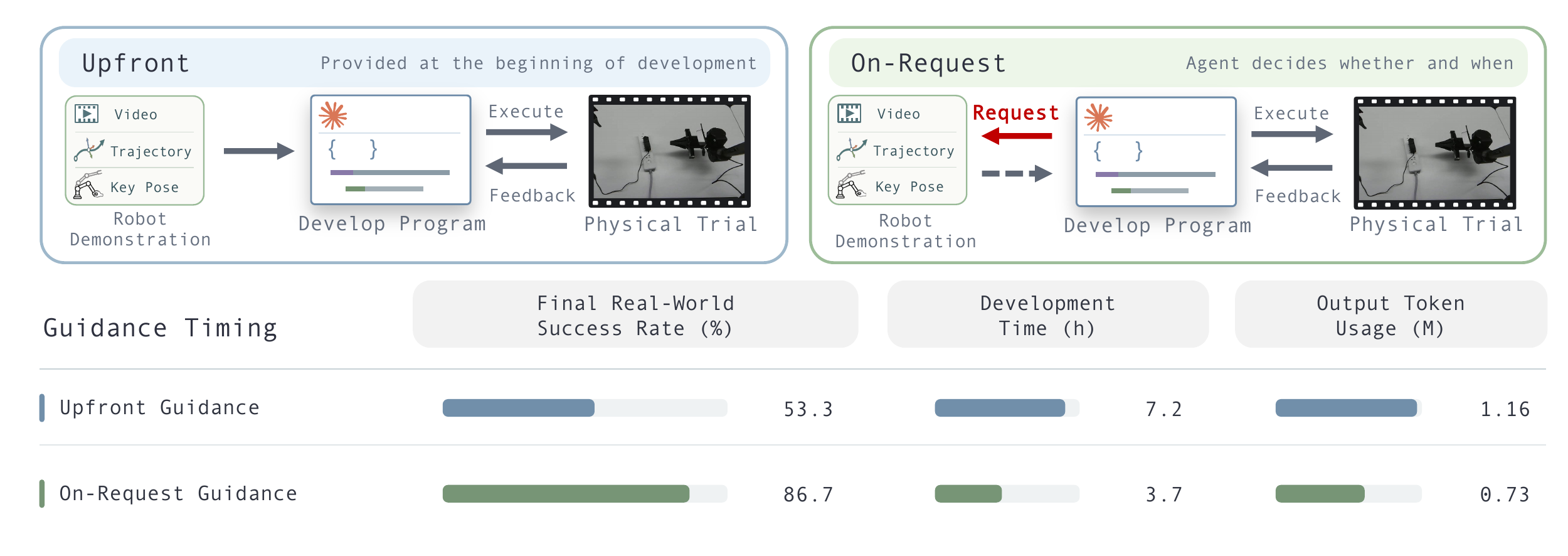}
    \caption{Guidance timing for charger insertion. Textual guidance and robot demonstrations are provided upfront or requested during real-only development. The lower panels report final success rate over 30 real-world trials of each frozen program, development time, and output token usage. On-Request Guidance shares the Robot Demonstrations condition and results from Figure~\ref{fig:investigation_guidance}.}
    \label{fig:investigation_guidance_timing}
\end{figure}

Human guidance supplies prior knowledge about manipulation strategies and their implementation~\cite{kang2026encore,cheng2026context}.
For charger insertion, the task goal leaves choices about grasping, plug orientation, and approach motion unresolved.
Demonstrations can supply a workable reference when execution feedback alone does not indicate how to revise the program~\cite{kang2026encore,liu2026rapid}.
We examine guidance form and timing using four conditions developed without simulation.
Figures~\ref{fig:investigation_guidance} and~\ref{fig:investigation_guidance_timing} summarize the protocols and results.

\paragraph*{Guidance Form.}
We compare three forms of assistance, each available when the agent requests it:
\begin{itemize}[leftmargin=1.5em,itemsep=0.3em,topsep=0.3em,parsep=0pt]
    \item \textbf{Textual Guidance.}
    The agent can request textual advice about manipulation strategies, action constraints, and corrections.
    No demonstration is provided in this condition.

    \item \textbf{Robot-Egocentric Human Video.}
    In addition to textual guidance, the agent can request videos of a person performing the task with their hands, recorded from the robot's execution viewpoint.

    \item \textbf{Robot Demonstrations.}
    In addition to textual guidance, the agent can request kinesthetic demonstrations in which a person physically guides the robot arms through the task.
    Each demonstration provides video, arm trajectories, and key joint poses capturing arm configurations at critical action stages.
\end{itemize}

Both demonstration conditions use the same camera, fixed centrally between the robot arms on an independent mount.
Here, \emph{robot-egocentric} denotes the shared viewpoint during demonstration and robot execution.

The Robot Demonstrations condition achieves the highest final success rate, at 86.7\% (26/30), compared with 73.3\% (22/30) for Robot-Egocentric Human Video and 60\% for Textual Guidance (Figure~\ref{fig:investigation_guidance}).
The two demonstration conditions have similar development costs: 3.7 and 3.8 hours, with 0.73 and 0.75 million output tokens, respectively.
Textual Guidance requires 5.9 hours and 0.86 million output tokens.
At similar development cost, the robot-demonstration condition achieves higher final success than human video.

\paragraph*{Why demonstrations help.}
Both demonstration conditions combine higher final success with lower development costs than text alone.
A visible example conveys action order and object interactions, complementing textual descriptions of intent and constraints~\cite{murray2024teaching,kang2026encore}.
Human video supplies such a strategy, but the agent must still translate human hand movements into robot gripper and arm motions.
Robot demonstrations provide the strategy in the target embodiment together with recorded robot motion.
The video shows what the actions accomplish; arm trajectories and key joint poses provide quantitative references for approach paths and arm configurations.
These sources jointly constrain the action sequence and motion parameters to implement in code, reducing the motion details that must be inferred from images alone~\cite{kang2026encore,cheng2026context}.

\paragraph*{Guidance Timing.}
Active demonstration learning treats when and what to request as part of the learning process~\cite{hou2024give}.
We examine this choice in code-policy development by comparing two provision protocols for textual guidance and robot demonstrations:
\begin{itemize}[leftmargin=1.5em,itemsep=0.3em,topsep=0.3em,parsep=0pt]
    \item \textbf{Upfront Guidance.}
    The agent receives textual guidance and robot demonstrations at the outset.

    \item \textbf{On-Request Guidance.}
    The agent decides whether and when to request assistance, either before its first physical attempt or during subsequent program revision.
\end{itemize}
On-Request Guidance is the Robot Demonstrations condition above and the Real-Only condition below; all three labels refer to the same development run and evaluation results.

\FloatBarrier

On-Request Guidance achieves 86.7\% (26/30) final success, compared with 53.3\% (16/30) for Upfront Guidance, while development time decreases from 7.2 to 3.7 hours and output token usage from 1.16 to 0.73 million (Figure~\ref{fig:investigation_guidance_timing}).
On-request access ties assistance to the agent's current uncertainty: an initial grasp can be demonstrated before execution, or an unsuccessful motion can motivate a targeted example or correction.
This lets guidance inform the next code revision as new difficulties emerge during physical testing.

\subsection{When and How Should Simulation Support Real-World CaP?}
\label{sec:investigation_simulation}

Simulation can serve as a development environment before transfer or as an on-demand tool for experiments during real-world development~\cite{hu2026simex}.
We examine these roles through three exploration protocols:
\begin{itemize}[leftmargin=1.5em,itemsep=0.3em,topsep=0.3em,parsep=0pt]
    \item \textbf{Real-Only.}
    The agent develops its program through real-world execution and feedback, without simulation.
    Textual guidance and robot demonstrations are available on request.

    \item \textbf{Sequential Real-Sim-Real.}
    The agent first receives robot demonstrations in the real environment and constructs a simulation based on the real setup.
    It then refines its program through ten rounds of simulated exploration in four parallel environments, the maximum supported by the available hardware.
    Each round ends when all four environments are reset together.
    The resulting program is then transferred to the robot for evaluation.

    \item \textbf{Simulation on Demand.}
    The agent develops its program within a real-world feedback loop and can use or construct a simulator to investigate a current development question.
    It decides whether and when to simulate, which task segment and initial state to examine, and which candidate program revisions to test.
    Simulated experiments can precede the first physical attempt or occur during later revision.
    Textual guidance and robot demonstrations remain available on request throughout development.
\end{itemize}

\paragraph*{Comparison structure.}
Real-Only and Simulation on Demand share on-request access to textual guidance and robot demonstrations and the same physical scene-reset budget.
Their comparison examines what simulated experimentation adds to real-world development.
Sequential Real-Sim-Real provides a complementary workflow baseline with upfront demonstrations and a separate simulation stage before transfer.
Its ten simulation rounds are counted separately from human-assisted scene resets for physical trials.

\begin{figure}[H]
    \centering
    \includegraphics[width=\linewidth]{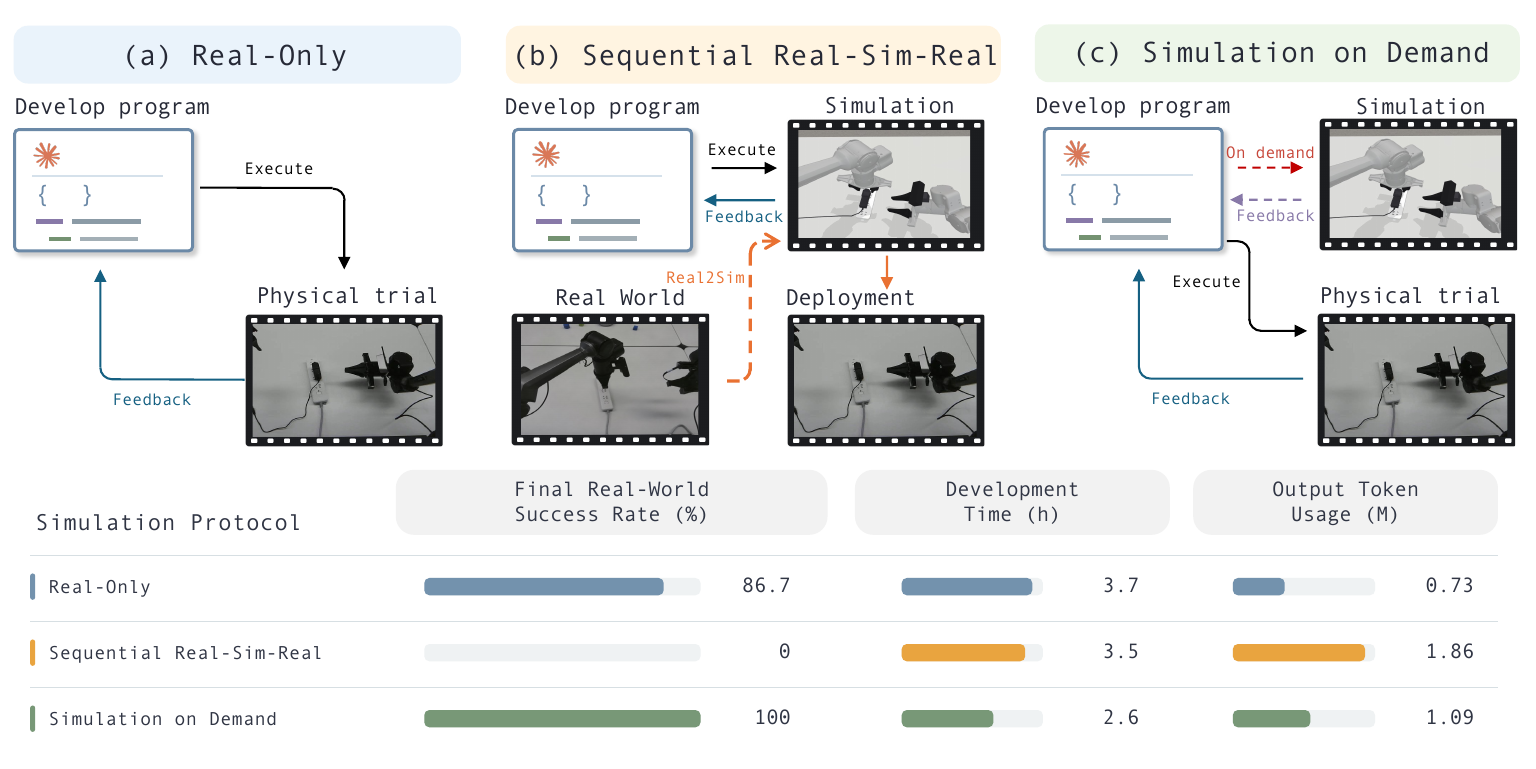}
    \caption{Simulation protocols and charger-insertion results. Top: (a) Real-Only uses physical feedback; (b) Sequential Real-Sim-Real uses ten simulation rounds before transfer, each ending with a joint reset of four parallel environments; (c) Simulation on Demand makes simulated experiments available within real-world development. Bottom: final success rate over 30 real-world trials of each frozen program, development time, and output token usage. Real-Only is the shared reference from Figures~\ref{fig:investigation_guidance} and~\ref{fig:investigation_guidance_timing}.}
    \label{fig:investigation_simulation}
    \label{fig:investigation_simulation_results}
\end{figure}
\FloatBarrier

\paragraph*{Results and development costs.}
Simulation on Demand achieves 100\% final real-world success, compared with 86.7\% for Real-Only, while development time decreases from 3.7 to 2.6 hours (Figure~\ref{fig:investigation_simulation_results}).
Output token usage increases from 0.73 to 1.09 million.

\paragraph*{Failure analysis.}
Sequential Real-Sim-Real achieves 0\% final real-world success after 3.5 hours of development and 1.86 million output tokens.
The transferred program fails at grasping the plug, before insertion.
In our setup, simulated images are cleaner, and simulated depth does not reproduce the real sensor's close-range blind regions.
The agent refines the image- and depth-dependent code directly against these simulated observations.
The failure is consistent with an observation mismatch: grasping decisions can depend on visual cues and depth measurements that are unreliable or unavailable on the robot~\cite{zhang2023close,li2024evaluating}.
The sequential workflow postpones testing these sensing assumptions on the robot until after simulated refinement.

\paragraph*{Implications.}
These results favor keeping robot demonstrations available on request and using simulation within real-world development.
Simulated experiments examine candidate revisions, while physical trials test the sensing and action assumptions on which those revisions depend~\cite{hu2026simex}.
This emphasis on interaction evidence also aligns with the diagnosis-driven use of offline priors advocated in online reinforcement learning~\cite{ma2026beyond}.
Together, the guidance and simulation studies motivate adaptive experience acquisition: the agent chooses what evidence to obtain as it revises the program.
Section~\ref{sec:recipe} develops this approach in a unified loop.

\FloatBarrier
\section{The ACE-CaP-0 Recipe}
\label{sec:recipe}

\begin{figure}[ht]
    \centering
    \includegraphics[width=0.99\linewidth]{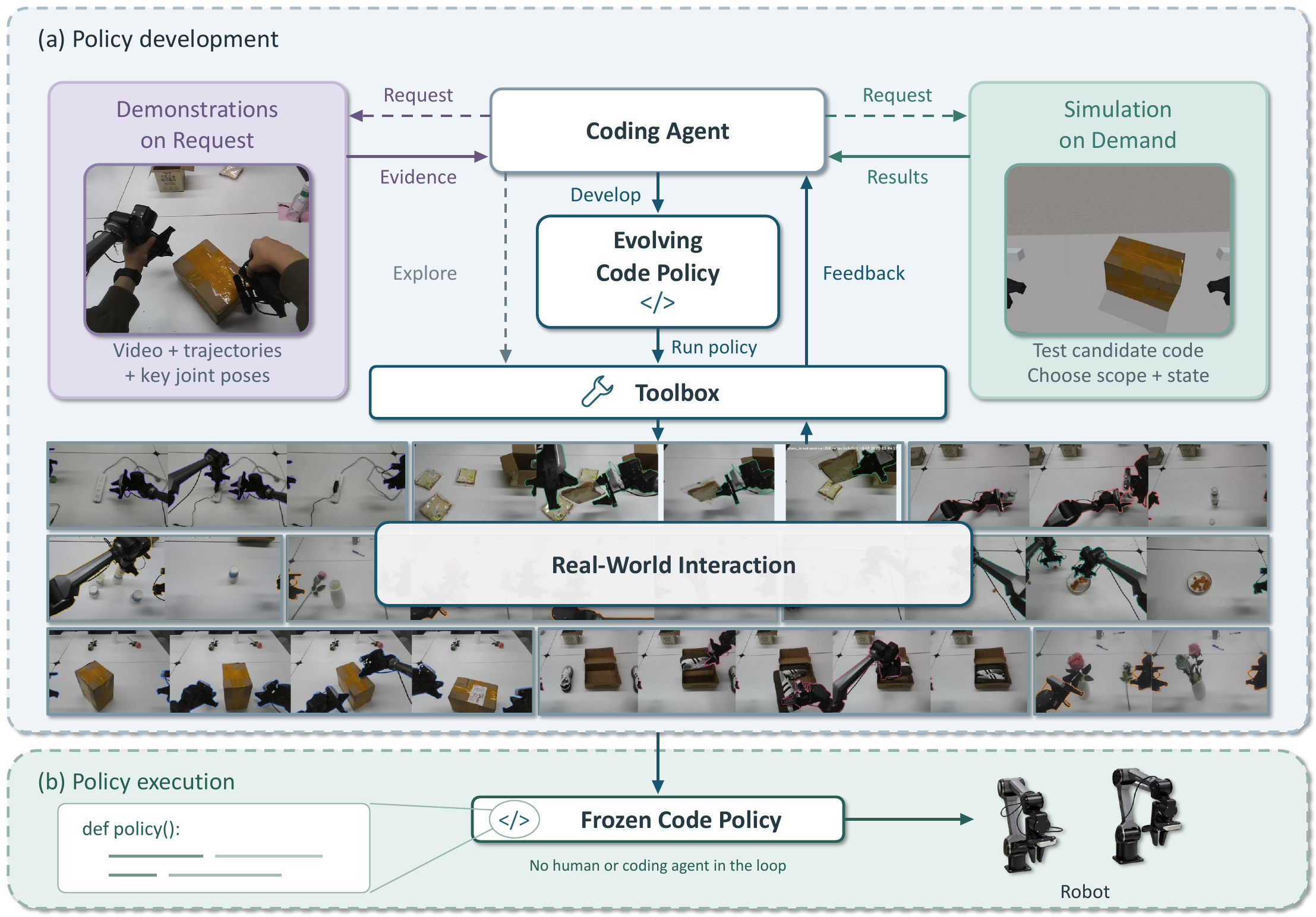}
    \caption{ACE-CaP-0: from policy development to standalone execution.
    (a) The coding agent writes and revises an executable policy using
    real-world feedback, with a shared robot toolbox that supports both policy
    execution and direct exploration by the agent.
    Throughout development, the agent can request kinesthetic demonstrations
    that provide video, arm trajectories, and key joint poses, or test
    candidate code in simulation for a chosen task segment and initial state.
    (b) The resulting code policy is frozen and executes through
    the robot's interfaces without a human or coding agent in the loop.
    Section~\ref{sec:real_world_evaluation} evaluates these frozen policies
    without further updates.}
    \label{fig:ace_cap_0_method}
\end{figure}

ACE-CaP-0 makes experience acquisition part of policy development.
The coding agent writes and revises an executable program, choosing when to request robot demonstrations, run simulated experiments, or test the program on the robot.
Building on the guidance and simulation choices examined in Section~\ref{sec:investigation}, this section describes how the resulting evidence informs program revision and how the final policy executes independently of the coding agent.
Figure~\ref{fig:ace_cap_0_method} shows these two stages.

\subsection{A Unified Policy Development Loop}
\label{sec:recipe_development}

\paragraph*{The program as the working artifact.}
The agent begins with a task instruction, observations of the workspace, and access to the robot's perception and control interfaces.
It constructs an executable program that specifies action sequences, geometric computations, motion parameters, and gripper commands.
The coding agent writes and revises this program during development; the program controls the robot through the available interfaces~\cite{liang2023code}.
ACE-CaP-0 builds on the write--execute--inspect--revise loop of embodied coding agents~\cite{fu2026cap}, with demonstrations and simulation available as additional sources of evidence.

\paragraph*{Choosing the next development step.}
At each step, the agent considers the current program, the evidence already available, and the question that remains unresolved.
It can revise the code directly, request a demonstration, organize a simulated experiment, or execute a candidate program on the robot.
A missing manipulation strategy can motivate a demonstration request; uncertainty about a candidate implementation can motivate a simulated comparison; physical execution tests how the program behaves in the real setup.
Demonstrations and simulation can be used before the first physical attempt or during later revision, and neither is required in every iteration.

\paragraph*{Using feedback to continue development.}
Physical outcomes inform both the next code revision and the choice of further experience.
A failure can expose an incorrect action sequence, geometric assumption, or motion parameter; it can also show that the current evidence is insufficient to identify a useful correction.
The sources can inform one another: demonstrations can suggest interactions to simulate, while inconclusive simulations can motivate demonstration requests.
Development is subject to the scene-reset budget in Sections~\ref{sec:investigation} and~\ref{sec:eval_setup}; the budget counts resets rather than autonomous executions or human demonstrations.

\subsection{Demonstrations on Request}
\label{sec:recipe_demonstrations}

\paragraph*{Requesting evidence for a manipulation decision.}
The agent requests a demonstration when it judges that observing how a person performs the task with the robot would help it construct or revise the policy~\cite{kang2026encore}.
Before an initial attempt, the unresolved issue may be how to establish an object interaction or coordinate the two arms.
After a failed attempt, the agent may understand where the program failed but lack a workable alternative for that action.
In the guidance protocol of Section~\ref{sec:investigation_human}, the agent also has access to textual suggestions about actions, constraints, or corrections.

\paragraph*{Reading object interaction and robot motion together.}
We use kinesthetic demonstrations in which a person physically guides the robot arms through the task.
Each demonstration provides video, recorded arm trajectories, and key joint poses that capture arm configurations at critical action stages.
These poses show the gripper placement needed to grasp the bread's plastic wrapper upon closure, the arm configurations for a shoe handover, and the plug--socket alignment for downward insertion.
The agent interprets these motion references together with the video to identify the configuration from which each action can proceed.

\paragraph*{Translating the demonstration into program changes.}
The demonstrated action order and the roles of the two arms inform the program's structure.
Approach directions, motion paths, and key joint poses inform the implementation of individual actions.
The agent interprets these references in relation to the current scene and the interfaces available to the policy, then writes or revises the corresponding code.
The resulting implementation still contains choices about geometry, timing, and contact whose physical consequences must be assessed.
The agent can examine a candidate implementation in simulation or test it on the robot, using the outcome to determine what requires further revision.

\subsection{Simulation on Demand}
\label{sec:recipe_simulation}

\paragraph*{Defining the experiment.}
Simulation on Demand lets the agent examine candidate program changes through simulated execution~\cite{hu2026simex}.
An experiment begins with a development question, such as whether an approach path avoids an obstruction, how a motion parameter affects the interaction, or how coordinated actions should be ordered.
The agent chooses the task segment, initial state, and candidate changes relevant to that question.
The scope can cover a single interaction or the complete task.
The initial state specifies which preceding task conditions the experiment takes as given.
The model must represent the geometry and interactions relevant to comparing the candidates.

\paragraph*{Constructing a test for the current question.}
The agent can use an existing simulator or construct and modify one for the experiment.
Workspace observations and previous physical outcomes inform the modeled geometry, state, and behavior.
Available demonstrations can further guide the choice of actions and interactions to examine.
The resulting simulator provides an executable setting in which the agent runs candidate programs and inspects their behavior.
The required correspondence with the physical system depends on the decision being made: a comparison of approach paths and a comparison of contact behaviors place different demands on the model~\cite{li2024evaluating}.
For code that uses images or depth, this includes whether the required observations are available and reliable on the robot.

\paragraph*{Taking simulated evidence back to the robot.}
The agent interprets the outcomes in relation to the question that motivated the experiment, then uses them to choose a code revision or the next physical test.
For example, a simulated path comparison can suggest an approach that avoids an obstruction while leaving the reliability of the eventual grasp to be checked on the robot.
The conclusion applies to the behavior and conditions examined by that experiment.
Physical execution then tests the revision on the robot.

If the real behavior differs in a way that affects the development decision, the agent reconsiders both the program and the assumptions of the simulated experiment.
It can revise the simulator, change the task segment or starting state, request a demonstration, or continue directly with real-world development.

\subsection{From Policy Development to Standalone Execution}
\label{sec:recipe_execution}
\label{sec:recipe_integration}

\paragraph*{Retaining experience in an executable policy.}
At the end of development, the agent produces a \emph{condensed code policy} that retains the task logic and execution details developed through interaction.
These choices are expressed through the program's action structure, geometric computations, and motion parameters.
We freeze the program before evaluation and assess it without further updates.

\paragraph*{Executing fixed rules with runtime feedback.}
During execution, the program calls the robot's perception and control interfaces.
Its decision rules remain fixed, while runtime observations can change the values computed by those rules and the resulting actions.
The perception and control components remain active, and the policy executes without a human or coding agent in the loop.
Section~\ref{sec:real_world_evaluation} measures the success rates and physical execution times of these programs across eight real-world manipulation tasks.

\clearpage
\section{Real-World Evaluation}
\label{sec:real_world_evaluation}

We evaluate ACE-CaP-0 on eight real-world manipulation tasks, measuring the success rate and physical execution time of the programs produced through the development process in Section~\ref{sec:recipe}.
Each task program is frozen before evaluation and executes through the robot's perception and control interfaces without a human or coding agent in the loop.

\subsection{Evaluation Setup}
\label{sec:eval_setup}

\begin{figure}[htbp]
    \centering
    \includegraphics[width=\linewidth]{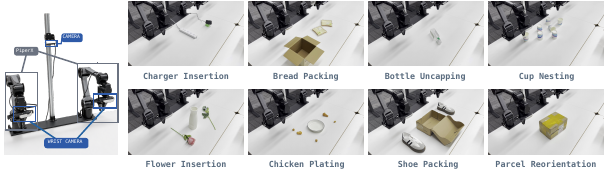}
    \caption{Real-world evaluation platform and tasks.
    Left: the dual-arm platform with one top-down camera and two wrist-mounted cameras.
    Right: representative scenes for the eight tasks, covering alignment, object reorientation, multi-object handling, and bimanual coordination.}
    \label{fig:exp_setting}
\end{figure}

\paragraph*{Platform and tasks.}
Our platform consists of two Agilex PIPERX robot arms and three Orbbec DaBai DC1 cameras, providing one top-down view and two wrist-mounted views (Figure~\ref{fig:exp_setting}).
The eight tasks span alignment, object reorientation, multi-object handling, and bimanual coordination.
They include inserting a charger into a socket, removing a bottle cap while stabilizing the bottle, and placing a pair of shoes into a box.
Table~\ref{tab:real_world_results} lists the objective of each task alongside its evaluation results.

\paragraph*{Policy development.}
We use Claude Opus 5.5~\cite{anthropic2026opus55} as the coding agent, with access to perception, reasoning, and robot control interfaces.
For each task, the agent develops an executable policy using the recipe in Section~\ref{sec:recipe}: it writes, executes, inspects, and revises the program, requesting robot demonstrations and conducting simulated experiments as needed.
Development allows at most ten human-assisted scene resets per task.
As in Section~\ref{sec:investigation}, resets used to collect demonstrations are excluded from this budget.
We measure output token usage by summing the tokens generated across all model calls during policy development, including demonstration interpretation, simulated experimentation, and program revision.

\paragraph*{Evaluation protocol.}
After development, we freeze each task program and evaluate it in 30 real-world trials, for 240 trials in total.
The same program is used throughout a task's evaluation, with runtime observations processed by its fixed decision rules.
Success rate is the fraction of trials in which the robot achieves the task objective.
We report mean physical execution time in minutes.
The evaluation trials are separate from those used for policy development.

\subsection{Reliable Execution for Real-World Manipulation}
\label{sec:real_world_results}

\begin{figure}[htbp]
    \centering
    \includegraphics[width=\linewidth]{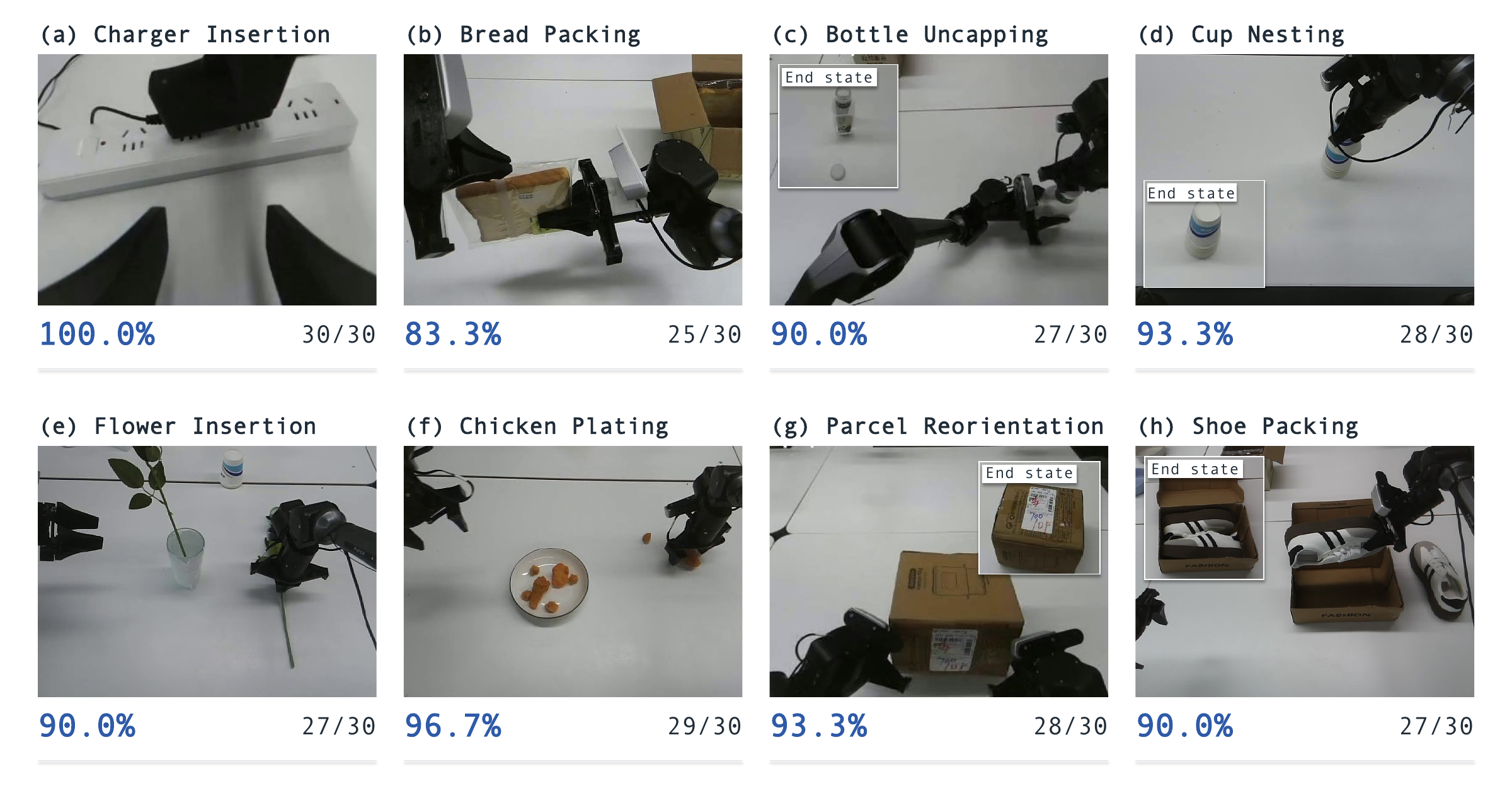}
    \caption{Real-world manipulation with the code policies developed by ACE-CaP-0.
    Each panel shows a task image and the success rate over 30 evaluation trials, with the successful trial count alongside it.
    Selected panels include end-state insets.}
    \label{fig:main_experiment}
\end{figure}

ACE-CaP-0 achieves an overall success rate of \textbf{92.1\%}, succeeding in \textbf{221 of 240 trials} across the eight tasks (Figure~\ref{fig:main_experiment} and Table~\ref{tab:real_world_results}).
Seven tasks reach success rates of at least 90\%.
\textit{Charger Insertion} succeeds in all 30 trials, while \textit{Chicken Plating} reaches 96.7\% (29/30).
\textit{Bread Packing} succeeds in 25 trials (83.3\%).

Mean physical execution times range from \textbf{0.97 to 2.27 minutes}, with six of the eight tasks averaging less than two minutes.
\textit{Parcel Reorientation} averages 0.97 minutes with 93.3\% success, while \textit{Charger Insertion} averages 1.65 minutes with 100\% success.
The two longest tasks, \textit{Bottle Uncapping} and \textit{Shoe Packing}, average 2.23 and 2.27 minutes, respectively, both with 90\% success.

The results demonstrate how ACE-CaP-0 turns experience acquired during policy development into reusable programs for reliable real-world manipulation.

\begin{table}[H]
    \centering
    \caption{Task objectives, execution performance, and output token usage.
    Each frozen program is evaluated in 30 trials; parentheses give successful and total trial counts.
    Time denotes mean physical execution time in minutes.
    Output token usage is the cumulative output across all model calls during development, in millions (M).}
    \label{tab:real_world_results}
    \small
    \renewcommand{\arraystretch}{1.18}
    \setlength{\tabcolsep}{4pt}
    \begin{tabularx}{\linewidth}{@{}l>{\raggedright\arraybackslash}Xrrr@{}}
        \toprule
        \textbf{Task} & \textbf{Task objective} & \textbf{Success rate} & \makecell[r]{\textbf{Mean time}\\\textbf{(min)}} & \makecell[r]{\textbf{Output token}\\\textbf{usage (M)}} \\
        \midrule
        Charger Insertion & Insert a charger plug into a socket. & 100.0\% (30/30) & 1.65 & 1.09 \\
        Bread Packing & Pack bread bags into a cardboard box. & 83.3\% (25/30) & 1.93 & 2.14 \\
        Bottle Uncapping & Stabilize a bottle and remove its cap. & 90.0\% (27/30) & 2.23 & 1.17 \\
        Cup Nesting & Insert multiple cups into one another. & 93.3\% (28/30) & 1.58 & 0.95 \\
        Flower Insertion & Insert flower stems into a vase. & 90.0\% (27/30) & 1.37 & 0.92 \\
        Chicken Plating & Plate the scattered chicken pieces. & 96.7\% (29/30) & 1.12 & 0.75 \\
        Parcel Reorientation & Rotate a parcel to the target orientation. & 93.3\% (28/30) & 0.97 & 1.07 \\
        Shoe Packing & Place a pair of shoes into a shoebox. & 90.0\% (27/30) & 2.27 & 1.87 \\
        \midrule
        \textbf{Overall} & & \textbf{92.1\% (221/240)} & --- & --- \\
        \bottomrule
    \end{tabularx}
\end{table}

\clearpage

\section{Related Work}
\label{sec:related_work}

We first review code-based robot control and learning from execution, including the practical constraints of physical trials.
We then examine how human guidance and simulated experience support policy development, focusing on the information they provide and when it is used.

\subsection{Code-as-Policies and Agentic Robot Learning}

\paragraph*{Code-based robot control.}
Foundation models support robot manipulation through both learned action policies and executable programs.
Vision-language-action models learn mappings from observations and instructions to actions~\cite{brohan2023rt,black2024pi_0}, while Code-as-Policies expresses behavior through programs that compose perception, geometric computation, and control interfaces~\cite{liang2023code}.
This representation makes task logic and motion parameters available for inspection and revision, but the capabilities of a code policy also depend on the interfaces it can call.
Studies across levels of abstraction show that high-level primitives can supply substantial task knowledge, while lower-level interfaces place more of the reasoning burden on the agent~\cite{fu2026cap}.
Code-based systems also differ in where decision-making resides: a generated program can govern execution, or an agent can remain in the loop to select tools and write code as the task unfolds~\cite{liang2023code,ge2026make,jia2026agent}.
The resulting behavior depends on both the available interfaces and the model's role in the control loop.
COAP studies fully code-driven execution through a shared stateful library, with its reported evaluation restricted to simulated RoboDojo tasks and public benchmark assets available during development~\cite{hu2026embodied}.
Experience-driven program development has also been evaluated on physical robots~\cite{huang2026scaling,hu2026simex}.
We study how human guidance and simulated experiments support policy development under real sensing and contact conditions.

\paragraph*{Robot learning from experience.}
Recent agentic systems use execution feedback to turn program generation into a process of policy improvement.
Execution traces support failure diagnosis and iterative repair, while validated solutions and repair knowledge can be retained as reusable skills~\cite{fu2026cap,lu2026aspire,ge2026make}.
Online exploration extends this process across tasks and embodiments~\cite{huang2026scaling}.
Execution outcomes can also be used to train code-generating models through reinforcement learning~\cite{fu2026cap,fan2026captain}.
Learned action policies can likewise improve through a combination of demonstrations, autonomous rollouts, and human interventions~\cite{intelligence2025pi}.
Systems that coordinate updates to an orchestrator and the learned policy it invokes further connect these mechanisms~\cite{zhang2026co}.
Program revision, skill reuse, and parameter learning offer complementary routes to improvement.

On physical robots, this learning process also depends on how experience is acquired and evaluated.
Improvements from retained experience need to be distinguished from those due to additional attempts or larger interaction budgets~\cite{wang2026sear}.
Real-world learning also requires execution, outcome verification, and scene reset; automating these operations is itself a substantial part of the learning system~\cite{xiao2026enpire,zhou2025autoeval}.
These constraints motivate our study of how additional experience can guide program revision when physical trials are scarce.

\subsection{Human and Simulated Experience for Policy Development}

\paragraph*{Demonstrations for policy development.}
Beyond providing training supervision, demonstrations can initialize exploration, condition a policy at deployment, or specify the behavior that a program should implement~\cite{bahl2022human,fu2024icrt,wang2023demo2code}.
In robot programming, language and visual examples help recover task structure, while richer demonstration records provide evidence for choosing and revising a manipulation strategy~\cite{murray2024teaching,kang2026encore}.
What the agent can infer depends on the information preserved in the demonstration: human videos convey object interactions and temporal structure, while robot demonstrations may pair visual evidence with recorded poses, actions, and gripper states~\cite{cheng2026context,kang2026encore}.
Studies of in-context adaptation show that action references and finer temporal resolution can improve performance for the models and tasks evaluated~\cite{cheng2026context,patel2026behaviorprompting}.
In code-policy development, the form of guidance shapes the strategies and motion details an agent can infer and test.

\paragraph*{Selective human assistance.}
Interactive robot learning also considers when and where human guidance is needed.
Confidence-based and interactive imitation methods request demonstrations or control interventions according to uncertainty, novelty, and risk~\cite{chernova2007confidence,hoque2021thriftydagger}.
Active demonstration learning also considers trajectory-level decisions about when to request an example and which example would be useful~\cite{hou2024give}.
For language-model agents, requests can resolve ambiguity in a plan, while planning-based demonstration collection invokes human assistance for task stages beyond the planner's supported behaviors~\cite{ren2023robots,sahoo2026tandem}.
Across these settings, the common principle is to seek assistance in response to the learner's current needs, with different query targets and learning objectives.
Our study examines this principle in program revision by comparing guidance provided at the outset with guidance requested by the agent as development proceeds.

\paragraph*{Simulation for policy development.}
Simulation complements physical interaction by making candidate behaviors available for repeated experimentation.
Dyna integrates modeled experience with learning and planning, while multi-fidelity policy search chooses between simulated and physical experiments according to their information value and cost~\cite{sutton1991dyna,marco2017virtual}.
Real-world feedback can also refine the simulator.
Real rollouts can update its parameters between policy-training rounds, while task-driven adaptation optimizes it for the resulting policy's real-world performance~\cite{chebotar2019closing,ren2023adaptsim}.
The usefulness of simulated evidence depends on the decision it supports and the errors that affect that decision; evaluations of sim--real agreement therefore examine policy rankings and behavior as well as simulated success~\cite{li2024evaluating}.

Agentic methods use simulation to support planning and program revision.
Language models can generate tasks and executable environments that provide opportunities for testing~\cite{wang2024gensim,aggarwal2026gym}.
Simulation can support action planning, including replanning from an updated real-world state after failure~\cite{liu2026simpact}, and the construction and verification of robot programs~\cite{kang2026coenv,li2026agentic}.
Demonstrations and physical trials can further shape what an agent tests.
RAPID derives task specifications, manipulation primitives, and verification environments from a human demonstration to guide iterative program synthesis~\cite{liu2026rapid}.
SimEX first develops a reusable toolbox through open-ended simulated experimentation, then uses limited physical trials to align the simulator, diagnose failures, and screen candidate repairs~\cite{hu2026simex}.

Together, these lines of work motivate examining how human guidance and simulated experimentation are organized during real-world policy development.
We investigate guidance modality and timing, and compare simulation-first development with allowing the agent to invoke or construct simulation during work on the real task.
In the latter setting, the agent chooses whether to simulate, which part of the task to model, which initial state to use, and which candidate changes to test.
The resulting recipe combines demonstrations on request and simulation on demand within a common code-policy development process.

\section{Conclusion}
\label{sec:conclusion}

ACE-CaP-0 makes experience acquisition part of real-world code-policy development.
Our charger-insertion investigation favors robot demonstrations available on request and simulated experiments targeted at current development questions.
Together, these findings suggest organizing experience around unresolved manipulation decisions, while checking the assumptions behind program revisions through physical execution.
The comparison with real-only development also reveals a resource trade-off: simulation on demand combines higher final success and shorter development time with greater output token usage.

Across eight tasks, the resulting frozen programs achieved 221 successes in 240 real-world evaluation trials (92.1\%).
These results show that skills acquired during development can persist in executable programs that respond to runtime observations, without a human or coding agent in the loop.
Their explicit task logic could support future improvements in execution efficiency, with task success revalidated after changes.

The investigation uses one development run per condition on one task; repeated evaluation measures program reliability, while consistency across independent development runs remains to be established.
Broader studies should include repeated development runs across tasks and test reuse under new conditions, with cost accounting that includes demonstration effort.
A central next question is how agents can estimate the value of additional experience for policy revision and weigh it against the human effort, computation, and robot time required.
This includes deciding when simulated evidence warrants a physical trial and when an unresolved manipulation choice calls for a demonstration.

\clearpage
\bibliographystyle{unsrtnat}
\bibliography{references}

\clearpage
\appendix
\section{Contributions and Author List}
\label{app:contributors}
\noindent\textbf{Core Contributors:} 
Changxu Wei\textsuperscript{1,2},
Yunpeng Qing\textsuperscript{1,3}, 
Guozheng Ma\textsuperscript{1,4},
\medskip

\noindent\textbf{Contributors:}
Bowen Liu\textsuperscript{1}, Zexin He\textsuperscript{1}, 
Yilun Kong\textsuperscript{1}, 
Haoming Gu\textsuperscript{1}, 
Jingyuan Liu\textsuperscript{1},
Tianyu Zhu\textsuperscript{1},
Jianfei Wang \textsuperscript{1},
Junqiang Kou\textsuperscript{1},
Huinan Mo \textsuperscript{1},
Hao Lv \textsuperscript{1},
Pengbin Li \textsuperscript{1},
Yuesong Wang \textsuperscript{1},
Bo Li\textsuperscript{1},
Huan Zhou\textsuperscript{1},
Shunyu Liu\textsuperscript{4},
Chunxiao Liu\textsuperscript{1},
Bixuan Jin\textsuperscript{1},
Shaopeng Yin\textsuperscript{1},
Fei Wang\textsuperscript{1},
\medskip

\noindent\textbf{Advisors:}

Zhi Hou\textsuperscript{1,$\dagger$},
Jiaxing Huang\textsuperscript{2},
Changqing Zou\textsuperscript{3},
Dacheng Tao\textsuperscript{4},
Xiaogang Wang\textsuperscript{1}.

\medskip
\textsuperscript{$\dagger$}Project Leader

\medskip
\noindent\textbf{Affiliations:}\\
\textsuperscript{1}ACE Robotics\\
\textsuperscript{2}The Hong Kong Polytechnic University\\
\textsuperscript{3}Zhejiang University \\
\textsuperscript{4}Nanyang Technological University \\


\medskip

\clearpage

\end{document}